\def\compileFigures{0}
\newcommand{\filename}{main}
\newcounter{figureNumber}

\documentclass{article}

\usepackage{microtype}
\usepackage{graphicx}
\usepackage{subcaption}
\usepackage{booktabs} 
\usepackage{hyperref}

\usepackage[accepted]{icml2025}

\usepackage{amsmath}
\usepackage{amssymb}
\usepackage{mathtools}
\usepackage{amsthm}
\usepackage{url}            
\usepackage{amsfonts}       

\usepackage[capitalize,noabbrev]{cleveref}

\usepackage{xcolor}         
\usepackage{multirow}

\usepackage[colorinlistoftodos]{todonotes}
\usepackage{tikz}
\usepackage{pgfplots}
\usepgfplotslibrary{fillbetween}
\usepackage{rotating}

\if\compileFigures1
\usetikzlibrary{external}
\usetikzlibrary{calc}
\usetikzlibrary{positioning}
\fi

\pgfplotsset{
    box plot/.style={
        /pgfplots/.cd,
        only marks,
        mark=-,
        mark size=1em,
        /pgfplots/error bars/.cd,
        y dir=plus,
        y explicit,
    },
    box plot box/.style={
        /pgfplots/error bars/draw error bar/.code 2 args={%
            \draw  ##1 -- ++(1em,0pt) |- ##2 -- ++(-1em,0pt) |- ##1 -- cycle;
        },
        /pgfplots/table/.cd,
        y index=2,
        y error expr={\thisrowno{3}-\thisrowno{2}},
        /pgfplots/box plot
    },
    box plot top whisker/.style={
        /pgfplots/error bars/draw error bar/.code 2 args={%
            \pgfkeysgetvalue{/pgfplots/error bars/error mark}%
            {\pgfplotserrorbarsmark}%
            \pgfkeysgetvalue{/pgfplots/error bars/error mark options}%
            {\pgfplotserrorbarsmarkopts}%
            \path ##1 -- ##2;
        },
        /pgfplots/table/.cd,
        y index=4,
        y error expr={\thisrowno{2}-\thisrowno{4}},
        /pgfplots/box plot
    },
    box plot bottom whisker/.style={
        /pgfplots/error bars/draw error bar/.code 2 args={%
            \pgfkeysgetvalue{/pgfplots/error bars/error mark}%
            {\pgfplotserrorbarsmark}%
            \pgfkeysgetvalue{/pgfplots/error bars/error mark options}%
            {\pgfplotserrorbarsmarkopts}%
            \path ##1 -- ##2;
        },
        /pgfplots/table/.cd,
        y index=5,
        y error expr={\thisrowno{3}-\thisrowno{5}},
        /pgfplots/box plot
    },
    box plot median/.style={
        /pgfplots/box plot
    }
}

\definecolor{matplotblue}{RGB}{31,119,180}
\definecolor{matplotorange}{RGB}{255,127,14}
\definecolor{matplotgreen}{RGB}{44,160,44}
\definecolor{matplotred}{RGB}{214,39,40}
\definecolor{matplotpurple}{RGB}{148,103,189}
\definecolor{matplotbrown}{RGB}{140,86,75}
\definecolor{matplotpink}{RGB}{227,119,194}
\definecolor{matplotgrey}{RGB}{127,127,127}
\definecolor{matplotbgreen}{RGB}{188,189,34}
\definecolor{matplotcyan}{RGB}{23,190,207}

\definecolor{col1}{RGB}{237,248,177}
\definecolor{col2}{RGB}{127,205,187}
\definecolor{col3}{RGB}{44,127,184}

\newcommand{\x}[0]{\mathbf{x}}

\newcommand{\IF}[0]{\mathcal{IF}}

\DeclarePairedDelimiterX{\infdivx}[2]{(}{)}{%
  #1\;\delimsize\|\;#2%
}
\newcommand{\infdiv}{KL\infdivx}

\begin{document}

\twocolumn[
\icmltitle{Machine Learning from Explanations}

\begin{icmlauthorlist}
\icmlauthor{Jiashu Tao}{sch}
\icmlauthor{Reza Shokri}{sch}
\end{icmlauthorlist}

\icmlaffiliation{sch}{School of Computing, National University of Singapore, Singapore, Singapore}

\icmlcorrespondingauthor{Jiashu Tao}{jiashut@comp.nus.edu.sg}
\icmlcorrespondingauthor{Reza Shokri}{reza@comp.nus.edu.sg}

\icmlkeywords{Machine Learning, ICML, explanation, interpretability}

\vskip 0.3in
]



\printAffiliationsAndNotice{}  

\begin{abstract}
Acquiring and training on large-scale labeled data can be impractical due to cost constraints. Additionally, the use of small training datasets can result in considerable variability in model outcomes, overfitting, and learning of spurious correlations. A crucial shortcoming of data labels is their lack of any reasoning behind a specific label assignment, causing models to learn any arbitrary classification rule as long as it aligns data with labels. To overcome these issues, we introduce an innovative approach for training reliable classification models on smaller datasets, by using simple explanation signals such as important input features from labeled data. Our method centers around a two-stage training cycle that alternates between enhancing model prediction accuracy and refining its attention to match the explanations. This instructs models to grasp the rationale behind label assignments during their learning phase. We demonstrate that our training cycle expedites the convergence towards more accurate and reliable models, particularly for small, class-imbalanced training data, or data with spurious features.
\end{abstract}

\section{Introduction}
Machine learning has excellent performance in many challenging tasks~\citep{dosovitskiy2020image, liu2021swin, zhang2020pushing}, reaching or even outperforming humans~\citep{silver2016mastering, alphacode} in controlled experiments. However, their real-life performance is often drastically worse, especially when there are natural \textit{spurious correlations} in training data~\citep{arjovsky2019invariant, ribeiro2016should}, and in so many cases where the \textit{training data set is small, heterogeneous, or unbalanced}.  All this makes it hard for the existing learning algorithms to learn the right set of robust rules that can generalize well from training data. This is a serious issue in many critical applications where machine learning promises to assist humans.  For example, the training data used in training human-level medical AI often contain spurious features, and models overly rely on spurious features to classify medical images, which results in untrustworthy diagnoses in practice~\citep{rieger2020interpretations}.  To make it worse, models tend to extract more spurious reasons from the smaller sets, making models more useless for minority groups~\citep{sagawa2020investigation}. If the model does not learn the true reason, it usually finds reasons that favor the majority in the training set because it is an empirical risk minimizer.  Even if the model up-weights the minority group to prevent them from being left out, which is a common practice, it cannot classify minorities well on unseen data because the reasons extracted cannot be generalized. This discrepancy in group performance is also the base of the algorithmic (un)fairness problem in machine learning.

Getting more data seems to help with some obvious problems, but with a significant cost (for obtaining a large amount of high-quality, accurately labeled data, and for training models on large datasets). More importantly, it does not necessarily lead to stable and generalizable models. We observe that running the same training algorithms on the same dataset can yield models of similar accuracies, but with drastically different decision functions and low prediction agreements among themselves~\citep{ross2017right, d2020underspecification, watson2022agree}.  This comes as no surprise, given the high dimensionality of input space and parameter space, for (classification) tasks with so few labels. To fully eliminate the ambiguity in parameter space, we need datasets of huge sizes, which prohibitively increases the cost of using machine learning for most real-world problems. Moreover, collecting more (class balanced) data is not always possible. For example, for (rare) disease diagnoses, there are not many new patients every year to expand the datasets. In manufacturing industries, defect detection systems for new products also do not have much data available. This hinders the use of distributionally robust optimization methods~\citep{sagawa2019distributionally, zhang2020coping, idrissi2022simple} where a class balanced validation set is required~\citep{kirichenko2022last}. This shows there is a need for better learning algorithms in small data regimes where standard training fails to identify the right reasons.

In this paper, \textbf{we show that guiding machine learning with simple explanations can significantly improve performance, reduce sample complexity, and increase stability, when the training dataset is too small, imbalanced, or contaminated by spurious cues for models with standard training to learn the right reasons}.  We assume that for some training data, besides labels, we have expert explanations for the assigned labels.  The explanation can have different degrees of complexity. In its simplest form, we assume that a small subset of input features is highlighted to contain the high-level reason for the assigned label. We propose an effective algorithm for machine learning with explanations, where we guide the model to identify the correct latent features most consistent with the explanations while optimizing the overall accuracy based on labels.  Our approach significantly outperforms the baselines in convergence rate, accuracy, robustness to spurious correlations, and stability (which influences generalizability), when data is insufficient. Under these settings, we show that baselines not only struggle under complex tasks, but also fail at properly learning the right classification reasons in extremely simple tasks (e.g., detecting geometric shapes). Thus, we argue that it is \textit{necessary} to incorporate explanations in learning algorithms if we aim at deploying trustworthy ML models that do not latch onto spurious or counter-intuitive signals. 

Some prior works have explored the idea of using prior knowledge to improve machine learning~\citep{ross2017right, rieger2020interpretations, Schramowski2020MakingDN, Shao2021RightFB}. Although their objective is to be ``right for the \textit{right} reasons'', these methods are actually penalizing models when they learn the \textit{wrong} reasons. As we show in our analysis, this does not necessarily result in learning the right reasons, thus it has a limited advantage to learning only from labels. This is also fundamentally different from our proposed approach. Conceptually, providing the explanation for right reasons is much easier than enumerating all possible ways that the models might make mistakes, and doing so during the data collection phase. If there are known spurious correlations, a more straightforward and more effective solution (compared to penalizing the model on learning the spurious correlations) would be to remove them and train models on clean data~\citep{friedrich2022typology}.

\section{Learning from explanations}
\subsection{Problem statement}
Given a labeled dataset $D=\{(\x, y)_i\}$, we also have access to explanations $e(\x)$, which are a subset or binary mask of the input features, of the label for each input point $\x$. The explanations are informative enough that they can sufficiently explain the labels. We want to train a model to produce outputs similar to the given labels and base the decisions using reasons close to the given explanations.

\subsection{Why do previous methods not work well?}
In this work, we focus on the image domain, where a wide range of model explanation methods have been studied. There are some prior works \citep{ross2017right, rieger2020interpretations, Schramowski2020MakingDN, Shao2021RightFB} on a similar problem to ours. Explanations in their settings are bounding boxes of either the main object or the spurious features, which differ from our definition of ``informative and sufficient subsets of input features''. Technically, they all adopt a loss-based approach, adding an explanation misalignment loss to the label loss:
\begin{equation} \label{eq:prior_optimization}
	\mathcal{L}_{\text{joint}} = \mathcal{L}_{\text{label}} + \lambda \mathcal{L}_{\text{expl}}.
\end{equation}

The additional loss is computed by taking the difference between the feature attributions $attr(\x)$ computed by certain model explanation methods and the given explanation $e(\x)$:
\begin{equation} \label{eq:prior_expl_loss}
	\mathcal{L}_{\text{expl}} = ||attr(\x)-e(\x)||.
\end{equation}

While it is tempting to reuse the existing algorithms for our problem, they do not lead to higher test accuracy than vanilla training in practice. The first reason is that using model explanations as a proxy of models' attention is unreliable. \citet{adebayo2018sanity} have shown that many popular saliency map based explanations do not even pass the sanity check. The second reason is with training. Optimization of the joint loss is often done via gradient descent. However, gradients of the two loss terms may point to different directions, creating a race condition that pulls and pushes the model into a bad local optimum. Imagine the two gradients counteract each other. The weights are then updated with negligible aggregated gradients. This leads to models converging more slowly or not even getting updated in the worst case. \citet{rieger2020interpretations} and \citet{friedrich2022typology} have empirically verified that these methods often do not outperform the vanilla models trained with labels only, and sometimes they are strictly worse than vanilla models.

\subsection{Tuning models' attention on latent features to learn from the explanation}
To design a working algorithm that teaches the reason to models, the first step is to teach models to identify the predictive features highlighted by explanations. However, as input pixels are less semantically meaningful for image data, we need to guide the model to recognize important features in the latent space. Since explanations are sufficient, the latent representations of the given explanations should also be distinctive enough for the classifier. Hence, our core idea is to make the latent features extracted by our models more similar to the those of the explanations. We minimize the difference between normalized latent features $\x_{\text{feat}}$ and $\x'_{\text{feat}}$ by optimizing the feature misalignment loss, which in our design is the KL divergence between two normalized feature maps using softmax:
\begin{equation} \label{eq:feat_loss}
	\mathcal{L}_{\text{feat}}(\x_\text{feat}, \x_\text{feat}') = \infdiv{\x'_\text{feat}}{\x_\text{feat}}
\end{equation}
There are two design choices in this loss function. The first one is using normalized feature maps. Without normalization, latent features can be very large or very small, potentially resulting in exploding or vanishing gradients. Secondly, we use KL divergence as the loss criterion, similar to the saliency guided training loss \cite{ismail2021improving}. The main reason is that we want the output distributions of feature extractors to be more similar, which can imply that the model's focus is primarily on the reasons region. Other loss functions may work empirically, but do not provide similar distributional intuition.

Since we also need to make models accurate, we minimize the label loss:
\begin{equation} \label{eq:ce_loss}
	\mathcal{L}_{\text{CE}}(y', y) = -y \log(y') - (1-y) \log(1-y').
\end{equation}

To avoid creating the same race condition mentioned in the last subsection, we propose to optimize them sequentially. Our training algorithm alternates between minimizing the label loss and the feature misalignment loss. \cref{algo:optimization} describes our two-stage optimization.

\begin{algorithm}[!b]
	\caption{Two-stage optimization}
	\label{algo:optimization}
	\let\IF\relax
	\begin{algorithmic}[1]
		\REQUIRE Input data $\x$, model $h=c\circ m \circ f$ consists of feature extractor $f$, mapping layer $m$, and fully connected layers $c$, target $y$, explanation $e(\x)$, learning rates $\eta_1$ and $\eta_2$ for cross entropy loss and feature map loss
		\STATE $\mathcal{L}_{\text{CE}} \gets -y \log(h(\x)) - (1-y) \log(1-h(\x))$
		\STATE $\theta_h \gets \theta_h - \eta_1\nabla_{\theta_h}\mathcal{L}_{\text{CE}}$
		\STATE $\x' \gets \x \otimes e(\x)$
		\STATE $\x'_\text{feat} \gets \text{softmax}(f(\x'))$
		\STATE $\x_\text{feat} \gets \text{softmax}(m(f(\x)))$
		\STATE $\mathcal{L}_{\text{feat}} \gets \infdiv{\x'_\text{feat}}{\x_\text{feat}}$
		\STATE $\theta_m \gets \theta_m - \eta_2\nabla_{\theta_m}\mathcal{L}_{\text{feat}}$
	\end{algorithmic}
\end{algorithm}

\begin{figure*}[!ht]
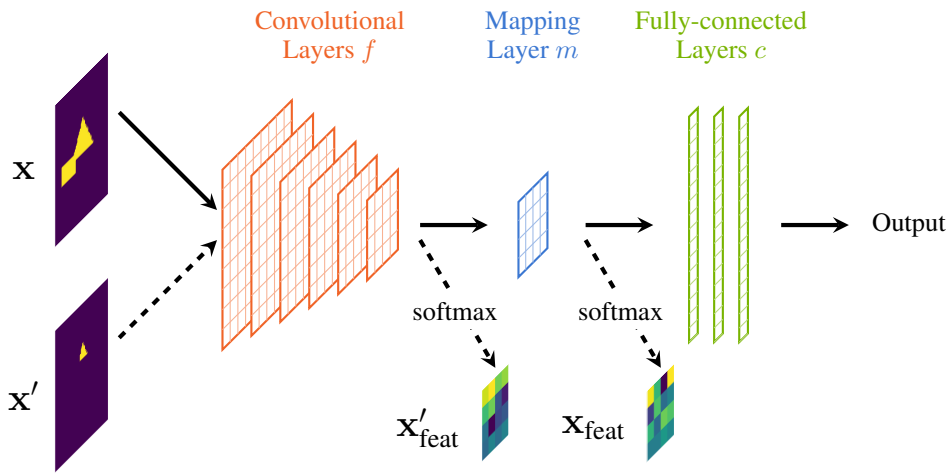

	\vskip 0.2in
	\begin{center}
	\if\compileFigures1
	\input{figure_scripts/pipeline}
	\else
	\centerline{\includegraphics[]{fig/\filename-figure\thefigureNumber.pdf}}
	\stepcounter{figureNumber}
	\fi
	\caption{A demonstration of how our proposed training pipeline uses the explanation. \textbf{Solid line:} First stage of the optimization, forward pass, and backpropagation of the cross entropy loss in the same way as the conventional machine learning training pipeline. In this example, an image $\x$ passes through the convolutional layers, which are the feature extractor, the self mapping layer, and the fully connected (FC) layers to reach the output. The cross entropy loss (or any other loss functions) is computed, and the gradients are backpropagated to \emph{all layers}. \textbf{Dotted line:} Second stage of the optimization, the new addition to the existing training pipeline that tunes the feature maps. In this diagram, the masked input $\x'$ based on the explanation is fed to the convolutional layers, and a feature map of $\x'_\text{feat}$ will be obtained (after applying softmax). We want the feature maps $\x_\text{feat}$ fed to the FC layers to contain as much information as $\x'_\text{feat}$. Hence we introduce a self mapping layer between the convolutional layers and the FC layers to learn a mapping function that filters out irrelevant information and extracts predictive features from the feature map. We compute the loss between the two feature maps and back-propagate the loss to the \emph{mapping layer}.}
	\label{fig:pipeline}
	\end{center}
	\vskip -0.2in
\end{figure*}

\cref{fig:pipeline} illustrates our two-stage training process. Given a convolutional neural network (CNN) $h$, we first decompose it into two parts: (i) the feature extractor $f$, which contains all the convolutional layers, and (ii) the classifier $c$, which contains all the fully connected (FC) layers. We insert a mapping layer $m$ in between $f$ and $c$. The mapping layer is a linear layer that maps 
features extracted by $f$ back to the same feature space. It will be trained to filter out irrelevant information and amplify signals from the reasons region. This addition is introduced because we do not want to update the weights of the feature extractor in the second stage. Otherwise, it would cause similar problems in prior work where two gradients counteract each other.

In the first stage, we do a normal forward pass and then backpropagate the label loss to update the entire model. In the second stage, we use the explanation $e(\x)$ as a binary mask over the input image $\x$ to obtain the masked input $\x'$. We can then pass the masked input to the feature extractor to obtain its feature map $\x'_\text{feat} = \text{softmax}(f(\x'))$. We then compute the feature map $\x_\text{feat} = \text{softmax}(m(f(\x)))$ and minimize its difference with the reasons' feature map $\x'_\text{feat}$ using \cref{eq:feat_loss}. We subsequently update the weights of the mapping layer $m$ by backpropagating the feature misalignment loss. More implementation details are available in \cref{app:implementation}. Ablation studies of our design choices can be found in \cref{app:ablation}.

\section{Empirical analysis}
We start our empirical analysis from simple geometric datasets where a human can apply a very simple rule to perfectly classify all images. However, such simple datasets are surprisingly challenging for machine learning models if they are trained with labels only. Not only do models fail to learn the simple rule from more than enough data, but they also often pick up wrong rules that do not generalize well. We have presented the intuition of our algorithm in the previous section, so we want to empirically verify that models trained with reasons using our training pipeline are better. After observing how conventional ML training algorithms fail on the easy geometric synthetic dataset, we naturally extend our analysis to the harder Bird dataset where domain knowledge or expert advice is needed to explain the decision labels. 

We want to explore the following problems in our empirical analysis:
\begin{enumerate}
	\item Do models benefit from learning with explanations? (\cref{sec:benefit_reasons})
	\item Do models learn the reasons suggested by the provided explanations? (\cref{sec:if_learn_reasons})
	\item Does training become more consistent if we present explanations to models? (\cref{sec:model_consistency})
	\item Can models still learn the suggested reason when there are multiple or spurious reasons for the same decision? (\cref{sec:multiple_reasons})
\end{enumerate}

\subsection{Datasets}
We focus on binary classification with three datasets in our empirical analysis. Two of the datasets are synthetic geometric datasets that are easy for humans but surprisingly challenging for machines, and one is a real dataset that is difficult even for normal humans. Details of the datasets are in \cref{app:datasets}.
\begin{itemize}
	\item \textbf{Triangle Orientation Dataset:} If a triangle is pointing upwards, it is assigned to be Class 1. Otherwise, it is Class 0. The explanation is a rectangular area highlighting the spatial arrangement of the vertex and the bottom.
	\item \textbf{Fox vs Cat Dataset:} Foxes (Class 1) have triangular heads and cats (Class 0) have round heads. The explanation is a square highlighting either the vertex or the arc. 
	\item \textbf{Bird Dataset:} We select Indigo Buntings and Blue Grosbeaks from the CUB-200-2011 dataset \citep{wah2011caltech} to form our binary Bird dataset.  The two species are visually identical except for the sizes of their beaks. The explanation is then a small square highlighting the beaks.
\end{itemize}

\subsection{Do models benefit from learning with explanations?} \label{sec:benefit_reasons}
\begin{figure*}[!t]
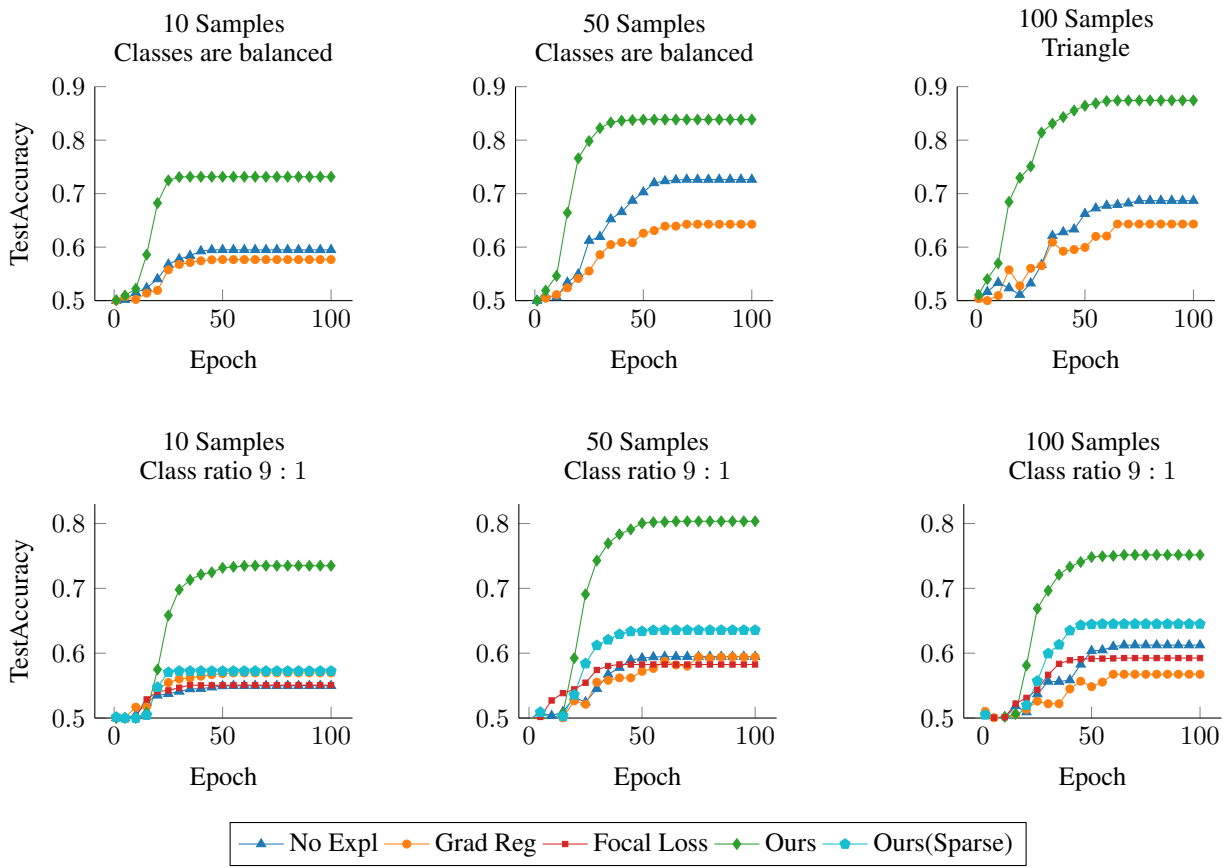

    \vskip 0.2in
	\begin{center}
     \begin{subfigure}[t]{0.31\textwidth}
     	\centering
     	\if\compileFigures1
\begin{tikzpicture}
	\pgfplotstableread[col sep = comma]{data/triangle_10_test_acc.csv}{\data}	
	\begin{axis}[scale=0.5,
		axis y line*=left,
		axis x line*=bottom,
		xlabel={Epoch},
		ymin=0.5,
		ymax=0.9,
		ylabel near ticks,
		ylabel={TestAccuracy},
		title style={align=center},
		title={10 Samples \\ Classes are balanced}
		]	
		\addplot[color=matplotblue, mark=triangle*, mark size=2pt] table [x={epochs}, y={no expl},] {\data};
		
		\addplot[color=matplotorange,mark=*, mark size=1.5pt] table [x={epochs}, y={grad},] {\data};
		
		\addplot[color=matplotgreen,mark=diamond*] table [x={epochs}, y={feat},] {\data};
	\end{axis}
\end{tikzpicture}
     	\else
     	\includegraphics[]{fig/\filename-figure\thefigureNumber.pdf}
     	\stepcounter{figureNumber}
     	\fi
     	\label{fig:triangle_10}
     \end{subfigure}
 	 \hfill
     \begin{subfigure}[t]{0.31\textwidth}
         \centering
         \if\compileFigures1
\begin{tikzpicture}
	\pgfplotstableread[col sep = comma]{data/triangle_50_test_acc.csv}{\data}	
	\begin{axis}[scale=0.5,
		axis y line*=left,
		axis x line*=bottom,
		xlabel={Epoch},
		ymin=0.5,
		ymax=0.9,
		title style={align=center},
		title={50 Samples \\ Classes are balanced}
		]	
		\addplot[color=matplotblue, mark=triangle*] table [x={epochs}, y={no expl},] {\data};
				
		\addplot[color=matplotorange,mark=*, mark size=1.5pt] table [x={epochs}, y={grad},] {\data};
				
		\addplot[color=matplotgreen,mark=diamond*] table [x={epochs}, y={feat},] {\data};
	\end{axis}
\end{tikzpicture}
         \else
         \includegraphics[]{fig/\filename-figure\thefigureNumber.pdf}
         \stepcounter{figureNumber}
         \fi
         \label{fig:triangle_50}
     \end{subfigure}
     \hfill
     \begin{subfigure}[t]{0.31\textwidth}
         \centering
         \if\compileFigures1
\begin{tikzpicture}
	\pgfplotstableread[col sep = comma]{data/triangle_100_test_acc.csv}{\data}	
	\begin{axis}[scale=0.5,
		axis y line*=left,
		axis x line*=bottom,
		xlabel={Epoch},
		ymin=0.5,
		ymax=0.9,
		title style={align=center},
		title={100 Samples \\ Triangle}
		]	
		\addplot[color=matplotblue, mark=triangle*, mark size=2pt] table [x={epochs}, y={no expl},] {\data};
		
		\addplot[color=matplotorange,mark=*, mark size=1.5pt] table [x={epochs}, y={grad},] {\data};
		
		\addplot[color=matplotgreen,mark=diamond*] table [x={epochs}, y={feat},] {\data};
	\end{axis}
\end{tikzpicture}
         \else
         \includegraphics[]{fig/\filename-figure\thefigureNumber.pdf}
         \stepcounter{figureNumber}
         \fi
         \label{fig:triangle_100}
     \end{subfigure}
 	 \hfill
 	 \vskip 0.2in
     \begin{subfigure}[t]{0.31\textwidth}
 	 	\centering
 	 	\if\compileFigures1
\begin{tikzpicture}
	\pgfplotstableread[col sep = comma]{data/triangle_10_skew01_test_acc.csv}{\data}	
	\begin{axis}[scale=0.5,
		axis y line*=left,
		axis x line*=bottom,
		xlabel={Epoch},
		ymin=0.5,
		ymax=0.83,
		ylabel near ticks,
		ylabel={TestAccuracy},
		title style={align=center},
		title={10 Samples \\ Class ratio $9:1$}
		]	
		\addplot[color=matplotblue, mark=triangle*, mark size=2pt] table [x={epochs}, y={no expl},] {\data};
		
		\addplot[color=matplotorange,mark=*, mark size=1.5pt] table [x={epochs}, y={grad},] {\data};
		
		\addplot[color=matplotgreen,mark=diamond*] table [x={epochs}, y={feat},] {\data};
		
		\addplot[color=matplotred,mark=square*, mark size=1pt] table [x={epochs}, y={focal},] {\data};
		
		\addplot[color=matplotcyan,mark=pentagon*] table [x={epochs}, y={sparse},] {\data};
	\end{axis}
\end{tikzpicture}
 	 	\else
 	 	\includegraphics[]{fig/\filename-figure\thefigureNumber.pdf}
 	 	\stepcounter{figureNumber}
 	 	\fi
 	 	\label{fig:triangle_10_skew01}
 	 \end{subfigure}
 	 \hfill
     \begin{subfigure}[t]{0.31\textwidth}
 	 	\centering
 	 	\if\compileFigures1
\begin{tikzpicture}
	\pgfplotstableread[col sep = comma]{data/triangle_50_skew01_test_acc.csv}{\data}	
	\begin{axis}[scale=0.5,
		axis y line*=left,
		axis x line*=bottom,
		xlabel={Epoch},
		ymin=0.5,
		ymax=0.83,
		title style={align=center},
		title={50 Samples \\ Class ratio $9:1$}
		]	
		\addplot[color=matplotblue, mark=triangle*] table [x={epochs}, y={no expl},] {\data};
				
		\addplot[color=matplotorange,mark=*, mark size=1.5pt] table [x={epochs}, y={grad},] {\data};
				
		\addplot[color=matplotgreen,mark=diamond*] table [x={epochs}, y={feat},] {\data};
		
		\addplot[color=matplotred,mark=square*, mark size=1pt] table [x={epochs}, y={focal},] {\data};
		
		\addplot[color=matplotcyan,mark=pentagon*] table [x={epochs}, y={sparse},] {\data};
	\end{axis}
\end{tikzpicture}
 	 	\else
 	 	\includegraphics[]{fig/\filename-figure\thefigureNumber.pdf}
 	 	\stepcounter{figureNumber}
 	 	\fi
 	 	\label{fig:triangle_50_skew01}
 	 \end{subfigure}
 	 \hfill
     \begin{subfigure}[t]{0.31\textwidth}
 	 	\centering
 	 	\if\compileFigures1
\begin{tikzpicture}
	\pgfplotstableread[col sep = comma]{data/triangle_100_skew01_test_acc.csv}{\data}	
	\begin{axis}[scale=0.5,
		axis y line*=left,
		axis x line*=bottom,
		xlabel={Epoch},
		ymin=0.5,
		ymax=0.83,
		title style={align=center},
		title={100 Samples \\ Class ratio $9:1$}
		]	
		\addplot[color=matplotblue, mark=triangle*, mark size=2pt] table [x={epochs}, y={no expl},] {\data};
		
		\addplot[color=matplotorange,mark=*, mark size=1.5pt] table [x={epochs}, y={grad},] {\data};
		
		\addplot[color=matplotgreen,mark=diamond*] table [x={epochs}, y={feat},] {\data};
		
		\addplot[color=matplotred,mark=square*, mark size=1pt] table [x={epochs}, y={focal},] {\data};
		
		\addplot[color=matplotcyan,mark=pentagon*] table [x={epochs}, y={sparse},] {\data};
	\end{axis}
\end{tikzpicture}
 	 	\else
 	 	\includegraphics[]{fig/\filename-figure\thefigureNumber.pdf}
 	 	\stepcounter{figureNumber}
 	 	\fi
 	 	\label{fig:triangle_100_skew01}
 	 \end{subfigure}
 	 \vskip 0.1in
  	 \if\compileFigures1
\begin{tikzpicture}
	\begin{axis}[scale=0.1,
		hide axis,
		xmin=0, xmax=1,
		ymin=0, ymax=1,
		legend columns=5,
		]
		\addlegendimage{color=matplotblue,mark=triangle*}
		\addlegendentry{No Expl};
		\addlegendimage{color=matplotorange,mark=*, mark size=1.5pt}
		\addlegendentry{Grad Reg};
		\addlegendimage{color=matplotred,mark=square*, mark size=1pt}
		\addlegendentry{Focal Loss};
		\addlegendimage{color=matplotgreen,mark=diamond*}
		\addlegendentry{Ours};
		\addlegendimage{color=matplotcyan,mark=pentagon*}
		\addlegendentry{Ours(Sparse)};
	\end{axis}
\end{tikzpicture}
  	 \else
  	 \includegraphics[]{fig/\filename-figure\thefigureNumber.pdf}
  	 \stepcounter{figureNumber}
  	 \fi
     \caption{Test performance on the Triangle Orientation dataset. \textbf{Top row:} Training with small sample sizes with balanced classes. \textbf{Bottom row:} Training with small sample sizes when positive data only make up $10\%$ of the training set.}
     \label{fig:triangle_dataset}
     \end{center}
	\vskip -0.2in
\end{figure*}

We train 30 models with early stopping for all datasets and sample sizes under each training setup. For balanced datasets, we use the vanilla training algorithm that trains with labels only. We also use the gradient regularization (Grad Reg) method proposed by \citet{ross2017right} as a baseline. For imbalanced datasets, we add two more settings. Firstly, we replace the cross entropy loss with focal loss \citep{lin2017focal}, a popular loss function for imbalanced datasets, to train models with labels only. Then we consider a sparse explanation setting where we only have explanations on the minority class. We train models with the normal cross entropy loss and tune their feature maps on points with explanations. The implementation details and hyper-parameters choices can be found in \cref{app:implementation}.

\begin{figure*}[t]
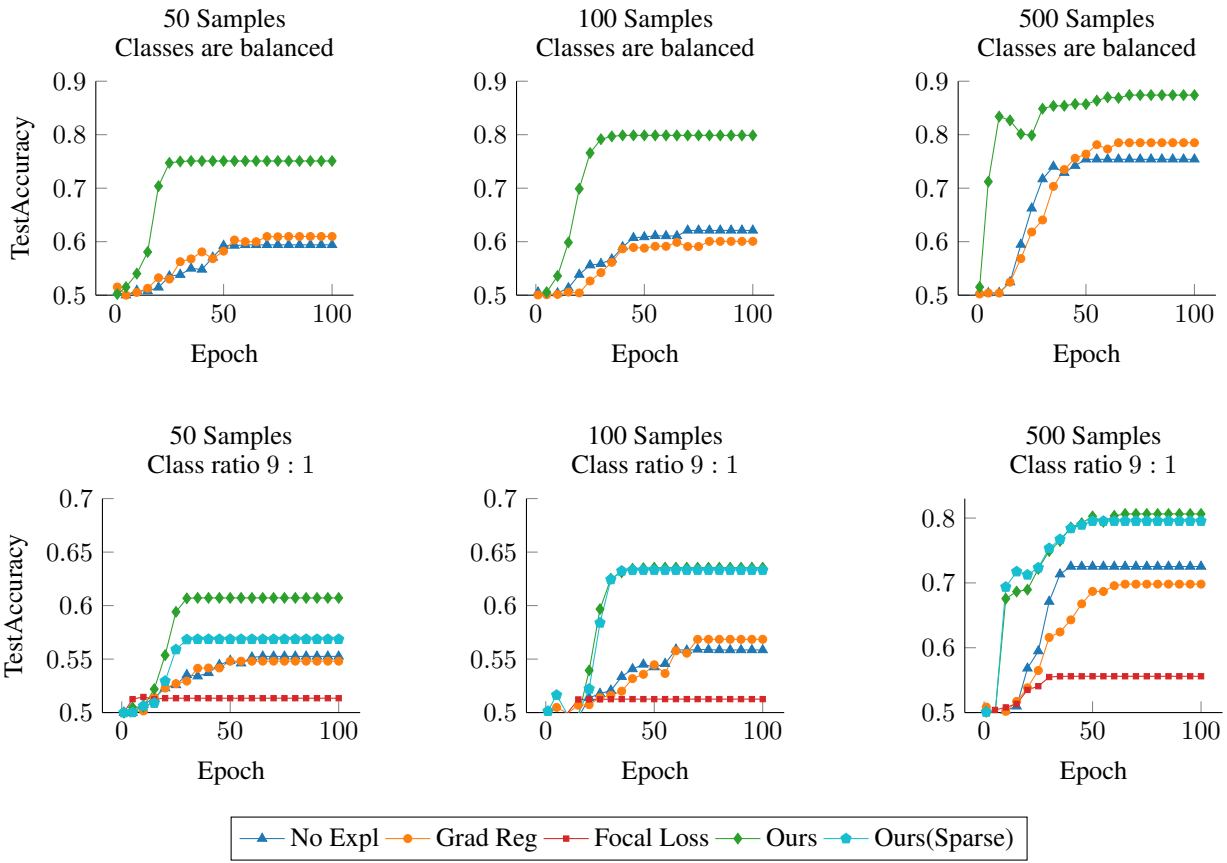

    \vskip 0.2in
	\begin{center}
     \begin{subfigure}[t]{0.31\textwidth}
     	\centering
     	\if\compileFigures1
\begin{tikzpicture}
	\pgfplotstableread[col sep = comma]{data/fox_50_test_acc.csv}{\data}	
	\begin{axis}[scale=0.5,
		axis y line*=left,
		axis x line*=bottom,
		xlabel={Epoch},
		ymin=0.5,
		ymax=0.9,
		ylabel near ticks,
		ylabel={TestAccuracy},
		title style={align=center},
		title={50 Samples \\ Classes are balanced}
		]	
		\addplot[color=matplotblue, mark=triangle*, mark size=2pt] table [x={epochs}, y={no expl},] {\data};
		
		\addplot[color=matplotorange,mark=*, mark size=1.5pt] table [x={epochs}, y={grad},] {\data};
		
		\addplot[color=matplotgreen,mark=diamond*] table [x={epochs}, y={feat},] {\data};
	\end{axis}
\end{tikzpicture}
     	\else
     	\includegraphics[]{fig/\filename-figure\thefigureNumber.pdf}
     	\stepcounter{figureNumber}
     	\fi
     	\label{fig:fox_50}
     \end{subfigure}
 	 \hfill
     \begin{subfigure}[t]{0.31\textwidth}
         \centering
         \if\compileFigures1
\begin{tikzpicture}
	\pgfplotstableread[col sep = comma]{data/fox_100_test_acc.csv}{\data}	
	\begin{axis}[scale=0.5,
		axis y line*=left,
		axis x line*=bottom,
		xlabel={Epoch},
		ymin=0.5,
		ymax=0.9,
		title style={align=center},
		title={100 Samples \\ Classes are balanced}
		]	
		\addplot[color=matplotblue, mark=triangle*, mark size=2pt] table [x={epochs}, y={no expl},] {\data};
		
		\addplot[color=matplotorange,mark=*, mark size=1.5pt] table [x={epochs}, y={grad},] {\data};
		
		\addplot[color=matplotgreen,mark=diamond*] table [x={epochs}, y={feat},] {\data};
	\end{axis}
\end{tikzpicture}
         \else
         \includegraphics[]{fig/\filename-figure\thefigureNumber.pdf}
         \stepcounter{figureNumber}
         \fi
         \label{fig:fox_100}
     \end{subfigure}
     \hfill
     \begin{subfigure}[t]{0.31\textwidth}
         \centering
         \if\compileFigures1
\begin{tikzpicture}
	\pgfplotstableread[col sep = comma]{data/fox_500_test_acc.csv}{\data}	
	\begin{axis}[scale=0.5,
		axis y line*=left,
		axis x line*=bottom,
		xlabel={Epoch},
		ymin=0.5,
		ymax=0.9,
		title style={align=center},
		title={500 Samples \\ Classes are balanced}
		]	
		\addplot[color=matplotblue, mark=triangle*] table [x={epochs}, y={no expl},] {\data};
				
		\addplot[color=matplotorange,mark=*, mark size=1.5pt] table [x={epochs}, y={grad},] {\data};
				
		\addplot[color=matplotgreen,mark=diamond*] table [x={epochs}, y={feat},] {\data};
	\end{axis}
\end{tikzpicture}
	     \else
	     \includegraphics[]{fig/\filename-figure\thefigureNumber.pdf}
	     \stepcounter{figureNumber}
	     \fi
         \label{fig:fox_500}
     \end{subfigure} 
 	\hfill
 	\vskip 0.2in
     \begin{subfigure}[t]{0.31\textwidth}
	\centering
	\if\compileFigures1
\begin{tikzpicture}
	\pgfplotstableread[col sep = comma]{data/fox_50_skew01_test_acc.csv}{\data}	
	\begin{axis}[scale=0.5,
		axis y line*=left,
		axis x line*=bottom,
		xlabel={Epoch},
		ymin=0.5,
		ymax=0.7,
		ylabel near ticks,
		ylabel={TestAccuracy},
		title style={align=center},
		title={50 Samples \\ Class ratio $9:1$}
		]	
		\addplot[color=matplotblue, mark=triangle*, mark size=2pt] table [x={epochs}, y={no expl},] {\data};
		
		\addplot[color=matplotorange,mark=*, mark size=1.5pt] table [x={epochs}, y={grad},] {\data};
		
		\addplot[color=matplotgreen,mark=diamond*] table [x={epochs}, y={feat},] {\data};
		
		\addplot[color=matplotred,mark=square*, mark size=1pt] table [x={epochs}, y={focal},] {\data};
		
		\addplot[color=matplotcyan,mark=pentagon*] table [x={epochs}, y={sparse},] {\data};
	\end{axis}
\end{tikzpicture}
	\else
	\includegraphics[]{fig/\filename-figure\thefigureNumber.pdf}
	\stepcounter{figureNumber}
	\fi
	\label{fig:fox_50_skew01}
 	\end{subfigure}
    \hfill
     \begin{subfigure}[t]{0.31\textwidth}
		\centering
		\if\compileFigures1
\begin{tikzpicture}
	\pgfplotstableread[col sep = comma]{data/fox_100_skew01_test_acc.csv}{\data}	
	\begin{axis}[scale=0.5,
		axis y line*=left,
		axis x line*=bottom,
		xlabel={Epoch},
		ymin=0.5,
		ymax=0.7,
		title style={align=center},
		title={100 Samples \\ Class ratio $9:1$}
		]	
		\addplot[color=matplotblue, mark=triangle*, mark size=2pt] table [x={epochs}, y={no expl},] {\data};
		
		\addplot[color=matplotorange,mark=*, mark size=1.5pt] table [x={epochs}, y={grad},] {\data};
		
		\addplot[color=matplotgreen,mark=diamond*] table [x={epochs}, y={feat},] {\data};
		
		\addplot[color=matplotred,mark=square*, mark size=1pt] table [x={epochs}, y={focal},] {\data};
		
		\addplot[color=matplotcyan,mark=pentagon*] table [x={epochs}, y={sparse},] {\data};
	\end{axis}
\end{tikzpicture}
		\else
		\includegraphics[]{fig/\filename-figure\thefigureNumber.pdf}
		\stepcounter{figureNumber}
		\fi
		\label{fig:fox_100_skew01}
	\end{subfigure}
	\hfill
     \begin{subfigure}[t]{0.31\textwidth}
		\centering
		\if\compileFigures1
\begin{tikzpicture}
	\pgfplotstableread[col sep = comma]{data/fox_500_skew01_test_acc.csv}{\data}	
	\begin{axis}[scale=0.5,
		axis y line*=left,
		axis x line*=bottom,
		xlabel={Epoch},
		ymin=0.5,
		ymax=0.83,
		title style={align=center},
		title={500 Samples \\ Class ratio $9:1$}
		]	
		\addplot[color=matplotblue, mark=triangle*] table [x={epochs}, y={no expl},] {\data};
				
		\addplot[color=matplotorange,mark=*, mark size=1.5pt] table [x={epochs}, y={grad},] {\data};
				
		\addplot[color=matplotgreen,mark=diamond*] table [x={epochs}, y={feat},] {\data};
		
		\addplot[color=matplotred,mark=square*, mark size=1pt] table [x={epochs}, y={focal},] {\data};
		
		\addplot[color=matplotcyan,mark=pentagon*] table [x={epochs}, y={sparse},] {\data};
	\end{axis}
\end{tikzpicture}
		\else
		\includegraphics[]{fig/\filename-figure\thefigureNumber.pdf}
		\stepcounter{figureNumber}
		\fi
		\label{fig:fox_500_skew01}
	\end{subfigure} 
	\vskip 0.1in
	\if\compileFigures1
\begin{tikzpicture}
	\begin{axis}[scale=0.1,
		hide axis,
		xmin=0, xmax=1,
		ymin=0, ymax=1,
		legend columns=5,
		]
		\addlegendimage{color=matplotblue,mark=triangle*}
		\addlegendentry{No Expl};
		\addlegendimage{color=matplotorange,mark=*, mark size=1.5pt}
		\addlegendentry{Grad Reg};
		\addlegendimage{color=matplotred,mark=square*, mark size=1pt}
		\addlegendentry{Focal Loss};
		\addlegendimage{color=matplotgreen,mark=diamond*}
		\addlegendentry{Ours};
		\addlegendimage{color=matplotcyan,mark=pentagon*}
		\addlegendentry{Ours(Sparse)};
	\end{axis}
\end{tikzpicture}
	\else
	\includegraphics[]{fig/\filename-figure\thefigureNumber.pdf}
	\stepcounter{figureNumber}
	\fi
	\caption{Test performance on the Fox vs Cat dataset. \textbf{Top row:} Training with small sample sizes with balanced classes. \textbf{Bottom row:} Training with small sample sizes when positive data only make up $10\%$ of the training set.}
	\label{fig:fox_dataset}
	\end{center}
	\vskip -0.2in
\end{figure*}

\begin{figure*}[!ht]
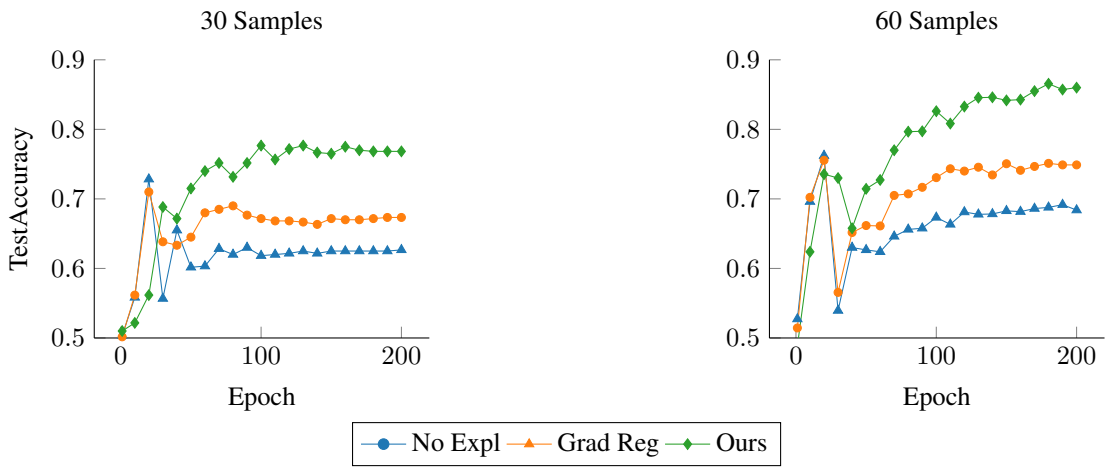

	\vskip 0.2in
	\begin{center}
	\centerline{
	\begin{subfigure}[t]{0.45\textwidth}
		\centering
		\if\compileFigures1
\begin{tikzpicture}
	\pgfplotstableread[col sep = comma]{data/bird_30_test_acc.csv}{\data}	
	\begin{axis}[scale=0.65,
		axis y line*=left,
		axis x line*=bottom,
		xlabel={Epoch},
		ymin=0.5,
		ymax=0.9,
		ylabel near ticks,
		ylabel={TestAccuracy},
		title={30 Samples}
		]	
		\addplot[color=matplotblue, mark=triangle*] table [x={epochs}, y={no expl},] {\data};
				
		\addplot[color=matplotorange,mark=*, mark size=1.5pt] table [x={epochs}, y={grad},] {\data};
				
		\addplot[color=matplotgreen,mark=diamond*] table [x={epochs}, y={feat},] {\data};
	\end{axis}
\end{tikzpicture}
		\else
		\includegraphics[]{fig/\filename-figure\thefigureNumber.pdf}
		\stepcounter{figureNumber}
		\fi
		\label{fig:bird_30}
	\end{subfigure} 
	\hfill
	\begin{subfigure}[t]{0.45\textwidth}
		\centering
		\if\compileFigures1
\begin{tikzpicture}
	\pgfplotstableread[col sep = comma]{data/bird_60_test_acc.csv}{\data}	
	\begin{axis}[scale=0.65,
		axis y line*=left,
		axis x line*=bottom,
		xlabel={Epoch},
		ymin=0.5,
		ymax=0.9,
		title={60 Samples}
		]	
		\addplot[color=matplotblue, mark=triangle*, mark size=2pt] table [x={epochs}, y={no expl},] {\data};
		
		\addplot[color=matplotorange,mark=*, mark size=1.5pt] table [x={epochs}, y={grad},] {\data};
		
		\addplot[color=matplotgreen,mark=diamond*] table [x={epochs}, y={feat},] {\data};
	\end{axis}
\end{tikzpicture}
		\else
		\includegraphics[]{fig/\filename-figure\thefigureNumber.pdf}
		\stepcounter{figureNumber}
		\fi
		\label{fig:bird_full}
	\end{subfigure}
	}
	\if\compileFigures1
\begin{tikzpicture}
	\begin{axis}[scale=0.01,
		hide axis,
		xmin=0, xmax=1,
		ymin=0, ymax=1,
		legend columns=3,
		]
		\addlegendimage{color=matplotblue,mark=*}
		\addlegendentry{No Expl};
		\addlegendimage{color=matplotorange,mark=triangle*}
		\addlegendentry{Grad Reg};
		\addlegendimage{color=matplotgreen,mark=diamond*}
		\addlegendentry{Ours};
	\end{axis}
\end{tikzpicture}
	\else
	\includegraphics[]{fig/\filename-figure\thefigureNumber.pdf}
	\stepcounter{figureNumber}
	\fi
    \caption{Test performance on the Bird dataset.}
    \label{fig:bird}
    \end{center}
	\vskip -0.2in
\end{figure*}

We observe from the top rows in \cref{fig:triangle_dataset} and \cref{fig:fox_dataset}, and \ref{fig:bird} that our training pipeline can accelerate the learning process and make final models generalize better on unseen data across all datasets when classes are \textbf{balanced} in the training set. Even when there is a \textbf{severe class imbalance}, where the class ratio in the training set is $9:1$, the bottom rows in \cref{fig:triangle_dataset} and \ref{fig:fox_dataset} show our models can reach higher test accuracy on a balanced test set. We report the standard deviations over 30 trials in \cref{table:std} in the appendix, and we observe our models are more consistent with standard deviations of one magnitude smaller.

To save annotation costs, sometimes it makes sense to only query explanations on data points from the \textbf{under-represented groups}. We simulate this realistic scenario by only training with explanations when the data are from the minority class (Class 1). The cyan lines in \cref{fig:triangle_dataset} and \ref{fig:fox_dataset} demonstrate that our training strategy still outperforms others, even with sparse explanations. This is very significant given the small sample sizes in experiments. For example, when we use a sample size of 50, the additional information only explains five images. Nevertheless, our algorithm can significantly improve the generalization performance and convergence speed. The reason for our method's usefulness in imbalanced settings without using popular re-weighting or sampling-based solutions shows that our models extract very useful features for minority groups under the guidance of explanations, so useful that the classification layers can easily distinguish the two classes in the latent space. 

In addition, \cref{fig:triangle_dataset}, \ref{fig:fox_dataset} and \ref{fig:bird} reveal that the existing algorithm from \citet{ross2017right} rarely outperforms the baseline, as the orange lines are often below the blue lines. This confirms that existing methods do not work well in practice.

\subsection{Do models learn the reasons suggested by the provided explanations?} \label{sec:if_learn_reasons}
While we have empirically observed that presenting explanations to models helps (\cref{sec:benefit_reasons}), we want to know if the superiority stems from the fact that our models learn the suggested reasons better. To analyze the alignment between the suggested reasons and the reasons models learned, previous work \citep{ross2017right, rieger2020interpretations, Shao2021RightFB, Schramowski2020MakingDN} relies mainly on model explanation methods to visualize saliency maps and inspecting the input features with high attribution values. However, as many researchers in the community have found out, many popular model explanation methods are bad proxies of models' behavior \citep{adebayo2018sanity, shah2021input}. Hence, to make our analysis more convincing, we construct additional test datasets that share the same classification rules as the synthetic datasets:
\begin{itemize}
	\item \textbf{Pentagon Orientation Dataset:} We construct this dataset by taking a vertical slice from the middle of each triangle in the test set of the Triangle Orientation dataset. The vertical slice is a pentagon. If the triangle points upwards, the pentagon should also point up.
	\item \textbf{Triangle vs Circle Dataset:} We construct this dataset by removing the rectangles in the test set of the Fox vs Cat dataset.
\end{itemize}

\begin{table*}[!htb]
	\caption{Average test accuracy of 30 models on test datasets sharing the same rules. For each row in the table, we report the test dataset we use, the training dataset, the size of the training dataset, the class ratio in the training dataset, and the test accuracy of different models obtained under different training schemes.}
	\label{table:same_reason_dataset}
	\vskip 0.15in
	\begin{center}
	\begin{small}
	\begin{sc}
		\begin{tabular}{cccccccc}
			\toprule
			\multicolumn{1}{c}{Test set} &\multicolumn{1}{c}{Train set} &\multicolumn{1}{c}{Size} &\multicolumn{1}{c}{Class ratio} &\multicolumn{1}{c}{No expl} &\multicolumn{1}{c}{Grad} &\multicolumn{1}{c}{Ours} &\multicolumn{1}{c}{Ours(sparse)}\\ 
			\midrule
			\multirow{8}{*}{Triangle vs Circle} & \multirow{8}{*}{Fox vs Cat}  & \multirow{2}{*}{50} & $1:1$ & 0.587 & 0.616 & \textbf{0.772} & \--- \\ 
			& & & $1:9$ & 0.561 & 0.549 & \textbf{0.618} & 0.580 \\ 
			& & \multirow{2}{*}{100} & $1:1$ & 0.623 & 0.606 & \textbf{0.813} & \--- \\ 
			& & & $1:9$ & 0.573 & 0.574 & \textbf{0.657} & 0.642 \\ 
			& & \multirow{2}{*}{500} & $1:1$ & 0.764 & 0.795 & \textbf{0.890} & \---\\ 
			& & & $1:9$ & 0.731 & 0.707 & \textbf{0.811} & 0.790\\
			& & \multirow{2}{*}{1000} & $1:1$ & 0.846 & 0.822 & \textbf{0.916} & \--- \\
			& & & $1:9$ &\textbf{0.834} & 0.824 & 0.825 & 0.824\\ 
			\midrule
			\multirow{6}{*}{Pentagon} & \multirow{6}{*}{Triangle} &
			\multirow{2}{*}{10} & $1:1$ & 0.561 & 0.565 & \textbf{0.710} & \--- \\ 
			& & & $1:9$ & 0.537 & 0.555 & \textbf{0.721} & 0.555 \\  
			& & \multirow{2}{*}{50} & $1:1$ & 0.713 & 0.596 & \textbf{0.833} & \--- \\ 
			& & & $1:9$ & 0.573 & 0.570 & \textbf{0.799} & 0.618 \\ 
			& & \multirow{2}{*}{100} & $1:1$ & 0.673 & 0.605 & \textbf{0.855} & \---\\ 
			& & & $1:9$ & 0.597 & 0.550 & \textbf{0.712} & 0.621\\ 
			\bottomrule
		\end{tabular}
	\end{sc}
	\end{small}
	\end{center}
	\vskip -0.1in
\end{table*}

If models have learned the suggested reasons from the training datasets, they should be able to apply the same reasons to the additional datasets and reach high accuracy. From the results listed in \cref{table:same_reason_dataset}, we observe that our models almost always perform the best on these datasets, suggesting our models have learned the most from the suggested reasons. Moreover, our approach outperforms the other training methods even in the sparse explanation setting. Hence, we safely claim that our models generalize better and faster because they grasp the suggested reasons better.

\subsection{Does training become more consistent if we present explanations to models?} \label{sec:model_consistency}
Without guidance from the reasons, models have too much freedom in choosing their decision functions. This often results in huge variances and discrepancies in models trained even with the same data. Since we have shown our models learn the given reasons better, it is important to see if they are more consistent with each other. We measure the pairwise agreement of final models by computing the percentage of identical predictions on test data. We then report the results in \cref{table:model_consistency}. Our models are always very consistent, implying they most probably use the same reasons for predictions. Models trained with labels only become more consistent when dataset size increases, supporting our hypothesis that these models can slowly infer the true reasons if the dataset is large enough.

\begin{table}[!ht]
	\caption{Average pairwise prediction agreement of 10 models when they are trained on the same dataset but with different strategies. Our models are always consistent, while models trained with labels only need large datasets to become consistent. }
	\label{table:model_consistency}
	\vskip 0.15in
	\begin{center}
	\begin{small}
	\begin{sc}
		\begin{tabular}{ccccc}
			\toprule
			\multicolumn{1}{c}{dataset} &\multicolumn{1}{c}{Size} &\multicolumn{1}{c}{No expl} &\multicolumn{1}{c}{Grad Reg} &\multicolumn{1}{c}{Ours}\\ 
			\midrule
			\multirow{3}{*}{Fox vs Cat} &  \multirow{1}{*}{50}  &  0.560  & 0.600 &  \textbf{0.927} \\ 
			& \multirow{1}{*}{100}  &  0.589  & 0.470   &  \textbf{0.924} \\ 
            &  \multirow{1}{*}{500}  &  0.782  & 0.707  &\textbf{0.944} \\
		    Triangle & 50 & 0.711 & 0.669 & \textbf{0.949} \\
			Bird	   & 	60	&	0.584	& 0.753		& \textbf{0.870} \\
			\bottomrule
		\end{tabular}
	\end{sc}
	\end{small}
	\end{center}
	\vskip -0.1in
\end{table}

\subsection{Can models still learn the suggested reason when there are multiple or spurious reasons for the same decision?} \label{sec:multiple_reasons}
\begin{figure*}[!htb]
	\vskip 0.2in
	\begin{center}
	\if\compileFigures1
\begin{tikzpicture}
	    \begin{axis}[
	    	every y tick label/.append style={font=\scriptsize},
	    	every x tick label/.append style={font=\scriptsize},
		width  = \textwidth,
		height = 4cm,
		major x tick style = transparent,
		ybar=2*\pgflinewidth,
		bar width=8pt,
		ymajorgrids, tick align=inside,
		major grid style={draw=white},
		ylabel = {TestAccuracy},
		symbolic x coords={
			Fox 50,Fox 100,Fox 500, Fox 1000, Triangle 10, Triangle 50, Triangle 100, Bird},
		xticklabel style={align=center, text width=1.5cm},
		xtick = data,
		scaled y ticks = false,
		enlarge y limits={value=.1,upper},
		enlarge x limits= true,
		ymin=0.5,
		ymax=0.85,
		legend cell align=left,
		legend style={
			at={(0.5,-0.2)},
			anchor=north,
			legend columns=-1,
			/tikz/every even column/.append style={column sep=0.5cm}
		}
		]
		\addplot[style={draw=none,fill=matplotblue,mark=none, fill opacity=0.5}]
		coordinates {
			(Fox 50, 0.620) 
			(Fox 100,0.570) 
			(Fox 500,0.611)
			(Fox 1000,0.601)
			(Triangle 10, 0.697)
			(Triangle 50, 0.694)
			(Triangle 100, 0.606)
			(Bird, 0.698)
		};
		
		\addplot[style={draw=none,fill=matplotorange,mark=none, fill opacity=0.5}]
		coordinates {
			(Fox 50,0.573) 
			(Fox 100,0.601) 
			(Fox 500, 0.561)
			(Fox 1000,0.605)
			(Triangle 10, 0.677)
			(Triangle 50, 0.600)
			(Triangle 100, 0.575)
			(Bird, 0.733)
		};
		
		\addplot[style={draw=none,fill=matplotgreen!90!black,mark=none}]
		coordinates {
			(Fox 50,0.729) 
			(Fox 100, 0.727) 
			(Fox 500,0.698)
			(Fox 1000,0.728)
			(Triangle 10, 0.726)
			(Triangle 50, 0.796)
			(Triangle 100, 0.784)
			(Bird, 0.878)
		};
		
		\legend{No Expl, Grad Reg, Ours}
	\end{axis}
\end{tikzpicture}
	\else
	\centerline{\includegraphics[]{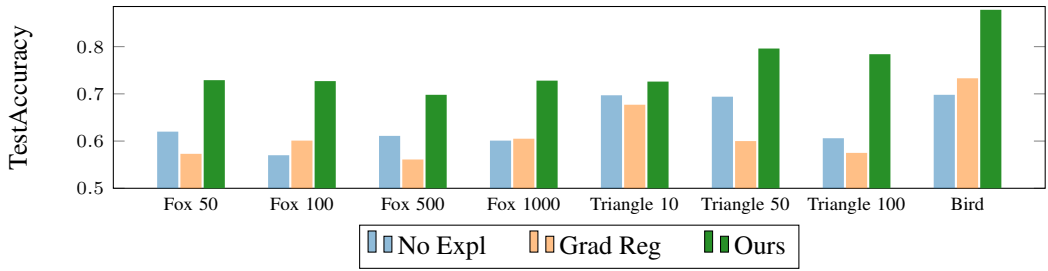}}
	\stepcounter{figureNumber}
	\fi
	\caption{Test accuracy on the clean test set when models are trained on a spurious training set. Our models consistently achieve the highest accuracy, implying that they have learned more from the suggested reasons.}
	\label{fig:spurious_clean}
	\end{center}
	\vskip -0.2in
\end{figure*}

To further demonstrate the usefulness of our method, we inject \textbf{spurious} features into training data. \cref{sec:spurious_dataset} details how we introduce the spurious features. If models can learn the right reasons to make decisions, training on spurious features should minimally affect the test performance on clean test sets. On the other hand, if models learn the spurious reasons in the training set, they will fail to generalize on the clean test set. \cref{fig:spurious_clean} shows our models are consistently the best-performing ones across all datasets and sample sizes on the clean test set, suggesting our models have learned the most out of the true reasons even when there are strong spurious signals. More results are provided in \cref{fig:spurious_same_reasons} and \cref{table:spurious_training} in the appendix.

\section{Discussion}
We show that our algorithm can improve machine learning models' convergence rate and generalization performance when the dataset is small. The advantage is clearer with imbalanced data, where querying explanations only for the minority group brings substantial improvement. We further verify that our models learn the right reasons better by testing models on alternative datasets that share the same classification rule. The higher consistency of our models also implies our models converge to learn the right reasons suggested by the given explanations, which subsequently explains the superior generalization performance. Lastly, by learning the right reasons, our models can be more robust against spurious features in the training data. With the reduced sample complexity, we provide a cost analysis in \cref{app:cost} to show curating explanations can be more cost effective than collecting more labeled data.
\section{Related work}
Previous work on making machine learning models learn desired reasons is mostly done via regularization with a given explanation. Most papers are from natural language processing (NLP), where the attention vector is a natural source of feature attribution. \citet{Pruthi2022EvaluatingEH} trained a student model to simulate the teacher model better using attention scores. \citet{Bao2018DerivingMA} utilized human attention on textual data to guide machine attention. \citet{Liu2019IncorporatingPW} treated feature attribution scores as priors to force the language models to produce similar feature attributions. Besides attention, \citet{chefer2022optimizing} regularized relevancy scores to shift transformers' focus to the foreground. However, some methods may only be useful in NLP due to the distinct characteristic of attention-based language models. Input pixels of images are also less semantically meaningful than words in language data. To our best knowledge, \citet{ross2017right} firstly explored the possibility of regularizing model explanation in training. However, their main point is that their algorithm can train diverse models focusing on different features with similar test accuracy. \citet{du2019learning} focused on the alignment between model explanations and given attribution scores by optimizing the difference and enforcing sparsity, but the resulting test accuracy did not improve. \citet{rieger2020interpretations} considered another form of explanation and focused on avoiding learning spurious correlations. Similarly, \citet{Shao2021RightFB} considered second-order derivatives as explanations and \citet{Schramowski2020MakingDN} used GradCAMs. However, these methods use the knowledge of ``wrong reasons'' as explanations, which is fundamentally different from our approach. They also all focus on large data regimes.
\section{Conclusion}
In this paper, we study a very important and useful machine learning setting: learning from explanations when training data is insufficient. We propose a novel two-stage optimization pipeline that teaches the model to extract features according to the given explanation. We demonstrate that our algorithm significantly outperforms the current training pipeline and previous methods in both synthetic and real datasets. We also show that our models can learn the suggested reasons better and apply the reasons to applicable alternative datasets. We think our method can naturally extend to multi-class classification, and we leave it as future work. One limitation is that we cannot make models to only extract and use the right reasons to make decisions. This is a common limitation of using a loss-based approach instead of constrained optimization. Unfortunately, real datasets that come with explanations are hard to find. Nonetheless, we hope our work can inspire more researchers to adopt this training pipeline so that more real-world datasets with explanations can be curated. In addition, we believe our method can also be used in model knowledge distillation where the teacher model provides both the labels and explanations (generated by a reliable model explanation technique) to train the student model.

As future work, we hope to extend our method to large language models (LLMs), where the explanations are the important words/tokens in the prompt, and one mapper layer is appended to each attention block in the style of LoRA~\cite{hu2022lora}. The second stage in the training pipeline would be similar to finetuning LLMs with LoRA in this way.
\section*{Impact Statement}
This paper presents work whose goal is to advance the field of Machine Learning. There are many potential societal consequences of our work, none which we feel must be specifically highlighted here.
\section*{Acknowledgement}
The authors would like to thank Martin Strobel for his input in the discussion and his help in creating the plots.

\bibliography{references}
\bibliographystyle{icml2025}


\newpage
\appendix
\onecolumn
\section{Datasets} \label{app:datasets}
We use two synthetic geometric datasets and one real datasets that involves expert domain knowledge in our paper.

\subsection{Triangle Orientation} \label{sec:triangle}
The first synthetic dataset is a binary classification dataset on the orientation of an triangle. A triangle is classified into class 1 if it is pointing upwards, otherwise 0. The explanation is the spatial arrangement of the vertex and the bottom line: if the vertex is above the bottom line, it is pointing upwards. The explanation is then represented as a mask highlighting the vertex and a portion of the bottom line. Each image is of grayscale with a size of $64\time 64$.

\subsection{Fox vs Cat} \label{sec:fox}
\begin{figure}[ht]
	\vskip 0.2in
	\begin{center}
	\begin{subfigure}[t]{0.23\textwidth}
		\centering
		\centerline{\includegraphics[width=\textwidth]{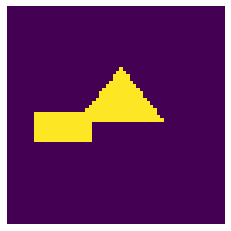}}
		\caption{Fox}
		\label{fig:fox}
	\end{subfigure}
	\hfill
	\begin{subfigure}[t]{0.23\textwidth}
		\centering
		\centerline{\includegraphics[width=\textwidth]{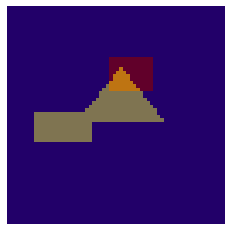}}
		\caption{Fox with a mask highlighting the vertex of its triangular head}
		\label{fig:fox_with_mask}
	\end{subfigure}
	\hfill
	\begin{subfigure}[t]{0.23\textwidth}
		\centering
		\centerline{\includegraphics[width=\textwidth]{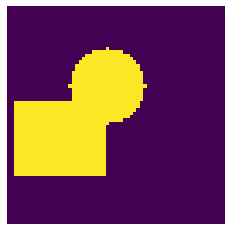}}
		\caption{Cat}
		\label{fig:cat}
	\end{subfigure}
	\hfill
	\begin{subfigure}[t]{0.23\textwidth}
		\centering
		\centerline{\includegraphics[width=\textwidth]{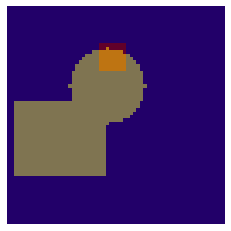}}
		\caption{Cat with a mask highlighting the arc of its round head}
		\label{fig:cat_with_mask}
	\end{subfigure}
	\caption{Fox (Left) vs Cat (Right).}
	\label{fig:fox_vs_cat}
	\end{center}
	\vskip -0.2in
\end{figure}

The second synthetic dataset is a minimalistic fox vs cat dataset. A fox has a triangular head and a rectangular body, while a cat has a circular head and a rectangular body. The explanation should focus on their heads. However, an explanation does not have to highlight the entire head. It only needs to highlight a distinctive feature of heads that sufficiently distinguishes a fox from a cat. Hence, the explanation for a fox is a mask highlighting the vertex of the triangular head, and the explanation for a cat is a mask highlighting the arc of the circular head (See \cref{fig:fox_vs_cat}). Each image is of grayscale with a size of $64\time 64$.

\subsection{Bird} \label{sec:bird}
\begin{figure*}[t]
     \vskip 0.2in
	\begin{center}
     \begin{subfigure}[t]{0.23\textwidth}
         \centering
         \includegraphics[width=\textwidth]{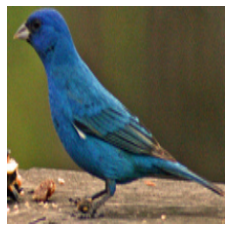}
         \caption{An Indigo Bunting}
         \label{fig:indigo_bunting}
     \end{subfigure}
     \hfill
     \begin{subfigure}[t]{0.23\textwidth}
         \centering
         \includegraphics[width=\textwidth]{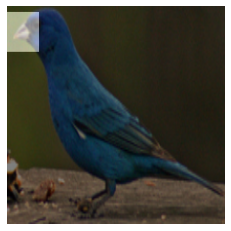}
         \caption{An Indigo Bunting with an explanation mask on its beak}
         \label{fig:indigo_bunting_with_mask}
     \end{subfigure}
     \hfill
     \begin{subfigure}[t]{0.23\textwidth}
         \centering
         \includegraphics[width=\textwidth]{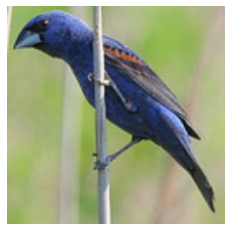}
         \caption{A Blue Grosbeak}
         \label{fig:blue_grosbeak}
     \end{subfigure}
     \hfill
     \begin{subfigure}[t]{0.23\textwidth}
         \centering
         \includegraphics[width=\textwidth]{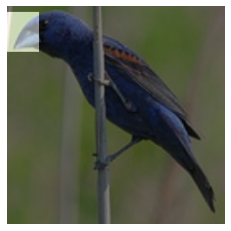}
         \caption{A Blue Grosbeak with an explanation mask on its beak}
         \label{fig:blue_grosbeak_with_mask}
     \end{subfigure}
     \caption{Indigo Buntings (Left) vs Blue Grosbeaks (Right). The difference is that Blue Grosbeaks have larger breaks. The explanations hence highlight their beaks.}
     \label{fig:bunting_vs_grosbeak}
     \end{center}
     \vskip -0.2in
\end{figure*}

We also extend our experiments to real datasets. We use a subset of the CUB-200-2011 (We will refer to it as the \textbf{Bird} dataset in the rest of the paper) dataset \citep{wah2011caltech}, a fine-grained classification dataset with photos of 200 species of birds. From the ground truth distribution of attributes for each individual class, we identify two species that are most similar to each other: the Indigo Bunting and the Blue Grosbeak (See \cref{fig:indigo_bunting} and \ref{fig:blue_grosbeak}). If we refer to their most likely attributes from the ground truth distribution, they are indistinguishable as they share the identical set of attributes. However, according to a bird watcher website \footnote{\url{https://www.sdakotabirds.com/diffids/blue_grosbeak_bunting.htm}}, Blue Grosbeaks have bigger and heavier beaks. Hence, the explanations for our Indigo Bunting vs Blue Grosbeak classification are masks highlighting the beaks if they are visible (See \cref{fig:indigo_bunting_with_mask} and \ref{fig:blue_grosbeak_with_mask}). We use the ground truth bounding box of birds to crop all images, before resizing them to the same size for training. 

\subsection{Injecting spurious features} \label{sec:spurious_dataset}
For the Fox vs Cat dataset, we add spurious features at fixed locations. For foxes, we add a square to the top left corner; for cats, we add a square to the top right corner. For the Triangle Orientation dataset, we add a square to a random location on the leftmost column for positive data. We add the square to a random location on the rightmost column for negative data. For the CUB bird dataset, we add a green square to a random place to the negative class, and add a red square to a random place to the positive class. We make sure that these spurious features do not block the reasons masks. However, they can be arbitrarily close to the reasons.
\begin{figure}[!t]
	\vskip 0.2in
	\begin{center}
	\begin{subfigure}[t]{0.31\textwidth}
		\centering
		\includegraphics[width=\textwidth]{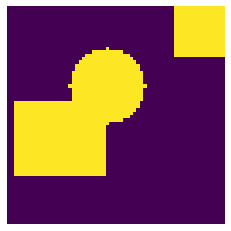}
		\caption{Spurious cat}
		\label{fig:cat_spurious}
	\end{subfigure}
	\hfill
	\begin{subfigure}[t]{0.31\textwidth}
		\centering
		\includegraphics[width=\textwidth]{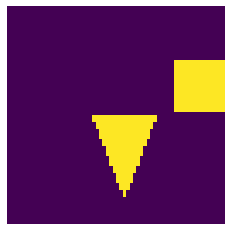}
		\caption{Spurious triangle pointing down}
		\label{fig:down_spurious}
	\end{subfigure} 
	\hfill
	\begin{subfigure}[t]{0.31\textwidth}
		\centering
		\includegraphics[width=\textwidth]{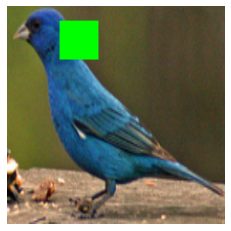}
		\caption{Spurious Indigo Bunting}
		\label{fig:bunting_spurious}
	\end{subfigure}
	\hfill
	\begin{subfigure}[t]{0.31\textwidth}
		\centering
		\includegraphics[width=\textwidth]{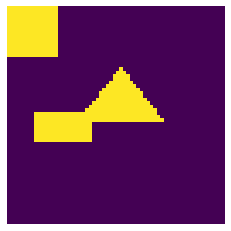}
		\caption{Spurious fox}
		\label{fig:fox_spurious}
	\end{subfigure}
	\hfill
	\begin{subfigure}[t]{0.31\textwidth}
		\centering
		\includegraphics[width=\textwidth]{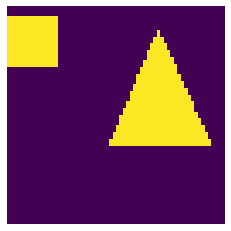}
		\caption{Spurious triangle pointing up}
		\label{fig:up_spurious}
	\end{subfigure}
	\hfill
	\begin{subfigure}[t]{0.31\textwidth}
		\centering
		\includegraphics[width=\textwidth]{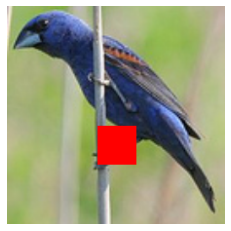}
		\caption{Spurious Blue Grosbeak}
		\label{fig:grosbeak_spurious}
	\end{subfigure} 
	\caption{\textbf{Top row:} images with spurious features injected from the negative class. \textbf{Bottom row:}  images with spurious features injected from the positive class. \textbf{Left:} a square is added to the top right corner for cat images and the top left corner for fox images. \textbf{Mid:} a square is added to the top left corner for triangles pointing down and the bottom left corner for triangles pointing up. \textbf{Right:} a green square is added to the lower right corner for indigo buntings, and a red square is added to the lower left corner for blue grosbeaks.}
	\label{fig:spurious_datasets}
	\end{center}
	\vskip -0.2in
\end{figure}

\section{Implementation} \label{app:implementation}
\subsection{Synthetic Datasets}
For the two synthetic datasets, we use a CNN with 3 convolutional layers and 1 FC layer. We use Adam with a learning rate of $0.01$ when we train with no explanations. When we add the gradient misalignment loss to the joint loss function, we optimize the joint loss with Adam with the same learning rate mentioned above. We set the balancing coefficient $\lambda$ of the gradient misalignment loss to be $0.001$ in \cref{eq:prior_optimization}, which keeps the two loss terms of similar magnitude to ensure maximum performance. This choice of hyper-parameter is mentioned by \citet{ross2017right} and also observed by us in hyper-parameter tuning. When using our method, we empirically evaluated the model with three learning rates for both $\eta_1$ and $\eta_2$: $\{0.1, 0.01, 0.001\}$. We then pick the best combination. Note that we did not tune these hyper-parameters to a great depth, so it is very likely our model can perform even better with more tuning. In summary, both optimizers are Adam with learning rates of $0.001$ when the class ratio is balanced with no spurious correlations. When there is a class imbalance or spurious correlation in data, we increase the learning rate $\eta_2$ of the feature loss optimizer to $0.01$. The empirical evidence that $\eta_2$ needs to be increased in these two scenarios is intuitive, because models need to update their focus more to generalize well to class balanced test set.

\subsection{Bird Dataset}
For the bird dataset, we use a simple CNN with 2 convolutional layers and 1 FC layer. This is simpler than the model used in synthetic datasets because the size of training set is smaller for this dataset. To reduce the chance of overfitting, we make the model simpler. We use SGD as the optimizer for this dataset. For the baseline method without explanations, the learning rate is $0.01$. For GradReg, the balancing coefficient is computed to make the two loss terms of similar magnitudes to ensure best performance. Similar to the hyper-parameter selection procedure explained above, we pick the best combination of $\eta_1$ and $\eta_2$ from empirical results. Across all settings, $\eta_1$ is always set to be $0.001$. When the class ratio is balanced with no spurious correlations, we set $\eta_2$ to be $0.001$. On the other hand, when there is a class imbalance or spurious correlation in data, we increase the learning rate $\eta_2$ of the feature loss optimizer to $0.01$.
\section{Ablation Studies} \label{app:ablation}
There are two novel designs in our algorithm:
\begin{enumerate}
	\item The mapping layer
	\item Two-stage optimization 
\end{enumerate}
We conduct ablation studies to see if any new designs bring significant improvement. \cref{table:ablation} shows that the combination of the two new designs yields the best final performance and fastest convergence rate, especially on the Bird dataset. With the mapping layer, the feature difference loss is only backpropagated to the mapping layer; otherwise it will be backpropagated to all convolutional layers. As argued before, updating the convolutional layers twice with different gradients may be less effective, and the two gradients might counteract each other. On the other hand, doing joint training is conceptually inferior to our two-stage training. The only difference would be that the feature difference loss in the joint optimization scheme will be computed before the feature extractor is updated by the label loss. This is less ideal because the gradient of feature difference loss might be outdated compared to the gradient in the two-stage training scheme, so the mapping layer will be less tuned. This inefficiency results in the slower convergence rate displayed in \cref{table:ablation}.

\begin{table*}[!ht]
	\caption{Model performance with standard deviation of two-stage training with mapping layer, two-stage training without mapping layer, and joint training with mapping layer. The numbers are computed at 20 epochs, {\color{red}{40} epochs} and {\color{blue}{convergence}} on synthetic datasets, and at 20 epochs,  {\color{red}{80} epochs} and {\color{blue}{convergence}} on Bird datasets because models take longer time to converge. }
	\label{table:ablation}
	\vskip 0.15in
	\begin{center}
	\begin{small}
	\begin{sc}	
		\begin{tabular}{ccccc}
			\toprule
			\multicolumn{1}{c}{Dataset} &\multicolumn{1}{c}{Size} &\multicolumn{1}{c}{Ours} &\multicolumn{1}{c}{No Mapping Layer} &\multicolumn{1}{c}{Joint Opt}
			\\ \midrule
			\multirow{12}{*}{Fox vs Cat} &  \multirow{3}{*}{50}  &  \pmb{$0.704 \pm 0.051$}  & $0.567 \pm 0.089$ &  {$0.632 \pm 0.099$} \\ 
			& & \color{red}\pmb{$0.751 \pm 0.029$} & \color{red}{$0.686 \pm 0.076$} & \color{red}{$0.749 \pm 0.051$} \\
			& & \color{blue}\pmb{$0.750 \pm 0.029$} & \color{blue}{$0.721 \pm 0.062$} & \color{blue}{$0.749 \pm 0.051$} \\
			\cmidrule{2-5}
			& \multirow{3}{*}{100}  & \pmb{$0.699 \pm 0.095$}  & $0.599 \pm 0.101$ &  {$0.683 \pm 0.105$} \\ 
			& & \color{red}\pmb{$0.799 \pm 0.019$} & \color{red}{$0.691 \pm 0.094$} & \color{red}{$0.783 \pm 0.071$} \\
			& & \color{blue}\pmb{$0.799 \pm 0.019$} & \color{blue}{$0.733 \pm 0.078$} & \color{blue}{$0.788 \pm 0.072$} \\
			\cmidrule{2-5}
			& \multirow{3}{*}{500}  & \pmb{$0.801 \pm 0.080$}  & $0.780 \pm 0.081$ &  {$0.605 \pm 0.155$} \\ 
			& & \color{red}\pmb{$0.854 \pm 0.030$} & \color{red}{$0.801 \pm 0.067$} & \color{red}{$0.760 \pm 0.159$} \\
			& & \color{blue}\pmb{$0.874 \pm 0.012$} & \color{blue}{$0.803 \pm 0.108$} & \color{blue}{$0.796 \pm 0.135$} \\
			\cmidrule{2-5}
			& \multirow{3}{*}{1000}  & \pmb{$0.738 \pm 0.158$}  & $0.698 \pm 0.155$ &  {$0.664 \pm 0.160$} \\ 
			& & \color{red}\pmb{$0.860 \pm 0.093$} & \color{red}{$0.797 \pm 0.167$} & \color{red}{$0.756 \pm 0.167$} \\
			& & \color{blue}{$0.903 \pm 0.017$} & \color{blue}\pmb{$0.909 \pm 0.016$} & \color{blue}{$0.834 \pm 0.034$} \\
			\midrule
			\multirow{9}{*}{Triangle} & \multirow{3}{*}{10}  & \pmb{$0.682 \pm 0.110$}  & $0.681 \pm 0.121$ &  {$0.557 \pm 0.089$} \\ 
			& & \color{red}\pmb{$0.732 \pm 0.082$} & \color{red}{$0.703 \pm 0.124$} & \color{red}{$0.627 \pm 0.112$} \\
			& & \color{blue}\pmb{$0.732 \pm 0.082$} & \color{blue}{$0.703 \pm 0.124$} & \color{blue}{$0.627 \pm 0.112$} \\
			\cmidrule{2-5}
			& \multirow{3}{*}{50}  & \pmb{$0.766 \pm 0.100$}  & $0.552 \pm 0.077$ &  {$0.753 \pm 0.111$} \\ 
			& & \color{red}\pmb{$0.836 \pm 0.026$} & \color{red}{$0.774 \pm 0.088$} & \color{red}{$0.827 \pm 0.060$} \\
			& & \color{blue}{$0.839 \pm 0.028$} & \color{blue}{$0.796 \pm 0.052$} & \color{blue}\pmb{$0.848 \pm 0.041$} \\
			\cmidrule{2-5}
			& \multirow{3}{*}{100}  & $0.730 \pm 0.115$ & $0.577 \pm 0.106$ & \pmb{$0.731 \pm 0.113$} \\
			& & \color{red}\pmb{$0.843 \pm 0.043$} & \color{red}{$0.778 \pm 0.089$} & \color{red}{$0.843 \pm 0.051$} \\
			& & \color{blue}\pmb{$0.874 \pm 0.018$} & \color{blue}{$0.801 \pm 0.076$} & \color{blue}{$0.871 \pm 0.044$} \\
			\midrule
			\multirow{6}{*}{Bird}	& \multirow{3}{*}{30} & \pmb{$0.562 \pm 0.088$}	& $0.495 \pm 0.053$		& {$0.500 \pm 0.007$} \\ 
			&  & \color{red}\pmb{$0.672 \pm 0.101$}	& \color{red}{$0.530 \pm 0.091$}		& \color{red}{$0.515 \pm 0.024$} \\
			&  & \color{blue}\pmb{$0.777 \pm 0.044$}	& \color{blue}{$0.522 \pm 0.038$}		& \color{red}{$0.545 \pm 0.057$} \\
			\cmidrule{2-5}
			& \multirow{3}{*}{60} & \pmb{$0.734 \pm 0.058$}	& $0.570 \pm 0.087$		& {$0.515 \pm 0.039$} \\ 
			&  & \color{red}\pmb{$0.797 \pm 0.085$}	& \color{red}{$0.692 \pm 0.069$}		& \color{red}{$0.613 \pm 0.089$} \\
			&  & \color{blue}\pmb{$0.860 \pm 0.079$}	& \color{blue}{$0.783 \pm 0.022$}		& \color{blue}{$0.772 \pm 0.039$} \\
			\bottomrule
		\end{tabular}
	\end{sc}
	\end{small}
	\end{center}
	\vskip -0.1in
\end{table*}
\section{Cost Analysis} \label{app:cost}
In this section, we analyze the benefit of training with explanations from the cost perspective. First of all, \cref{table:cost} shows that our method leads to lower sample complexity as the dataset size needed to reach a certain performance threshold is much smaller than that if using labels only. The table also shows that our method can achieve at least a 3x reduction in sample sizes, sometimes as much as 10x if the sample size is very small. It takes an Amazon Turk worker 1.2 cents \footnote{Numbers taken from \url{https://aws.amazon.com/sagemaker/data-labeling/pricing/}.} to label an image, and 3.6 cents to provide bounding boxes. The explanation used in our method is similar to bounding boxes where an annotator is only required to highlight an important region in the image. Hence, the cost to obtain explanations for each image would be approximately 3.6 cents, three times the cost of labels. However, considering how much reduction in dataset sizes our methods can achieve, we believe it is more cost-efficient to query for both labels and explanations than simply collecting more data to train models if model creators start from a set of unlabeled data.

\begin{table}[!ht]
	\caption{Test performance of models trained with no explanations and our method under different training set sizes.}
	\label{table:cost}
	\vskip 0.15in
	\begin{center}
	\begin{small}
	\begin{sc}	
		\begin{tabular}{ccccc}
		\toprule
			\multicolumn{1}{c}{Dataset} &\multicolumn{1}{c}{Size} &\multicolumn{1}{c}{No expl} &\multicolumn{1}{c}{Ours} \\ 
			\midrule
			\multirow{6}{*}{Fox vs Cat} &  50  &  0.594  &0.750 \\ 
			& 100  & 0.621  & 0.799 \\ 
			& 300  & 0.707  & 0.848 \\
			& 500  & 0.754 & 0.874 \\                         
			& 1000  & 0.837 & 0.903\\
			& 1500  & 0.881 & 0.920 \\
			\midrule
			\multirow{5}{*}{Triangle} & 10  & 0.595  & 0.732 \\ 
			& 50 & 0.726 & 0.839 \\ 
			& 100  & 0.687 & 0.874 \\
			& 200  & 0.827  & 0.900 \\
			& 500  & 0.863 & 0.913 \\
			\bottomrule
		\end{tabular}
	\end{sc}
	\end{small}
	\end{center}
	\vskip -0.1in
\end{table}
\section{Additional Results}
\begin{figure*}[!ht]
	\vskip 0.2in
	\begin{center}
	\if\compileFigures1
\begin{tikzpicture}
	    \begin{axis}[
	    	every y tick label/.append style={font=\scriptsize},
	    	every x tick label/.append style={font=\scriptsize},
		width  = \textwidth,
		height = 6cm,
		major x tick style = transparent,
		major grid style={draw=white},
		ybar=2*\pgflinewidth,
		bar width=10pt,
		ymajorgrids, tick align=inside,
		major grid style={draw=white},
		ylabel = {TestAccuracy},
		symbolic x coords={
			Fox 50,Fox 100,Fox 500, Fox 1000, Triangle 10, Triangle 50, Triangle 100},
		yticklabel style={align=center, text width=1.5cm},
		xtick = data,
		scaled y ticks = false,
		enlarge y limits={value=.1,upper},
		enlarge x limits= true,
		ymin=0.5,
		ymax=0.8,
		legend cell align=left,
		legend style={
			at={(0.5,-0.2)},
			anchor=north,
			legend columns=-1,
			/tikz/every even column/.append style={column sep=0.5cm}
		}
		]
		\addplot [style={draw=none,fill=matplotblue,mark=none, fill opacity=0.5}]
		coordinates {
			(Fox 50, 0.642) 
			(Fox 100,0.601) 
			(Fox 500,0.612)
			(Fox 1000,0.558)
			(Triangle 10, 0.698)
			(Triangle 50, 0.671)
			(Triangle 100, 0.603)
		};
		
		\addplot[style={draw=none,fill=matplotorange,mark=none, fill opacity=0.6}]
		coordinates {
			(Fox 50,0.563) 
			(Fox 100,0.613) 
			(Fox 500, 0.571)
			(Fox 1000,0.618)
			(Triangle 10, 0.691)
			(Triangle 50, 0.574)
			(Triangle 100, 0.555)
		};
		
		\addplot[style={draw=none,fill=matplotgreen!90!black,mark=none}]
		coordinates {
			(Fox 50,0.760) 
			(Fox 100,0.751) 
			(Fox 500,0.708)
			(Fox 1000,0.693)
			(Triangle 10, 0.711)
			(Triangle 50, 0.776)
			(Triangle 100, 0.780)
		};
		
		\legend{No Expl, Grad Reg, Ours}
	\end{axis}
\end{tikzpicture}
	\else
	\centerline{\includegraphics[]{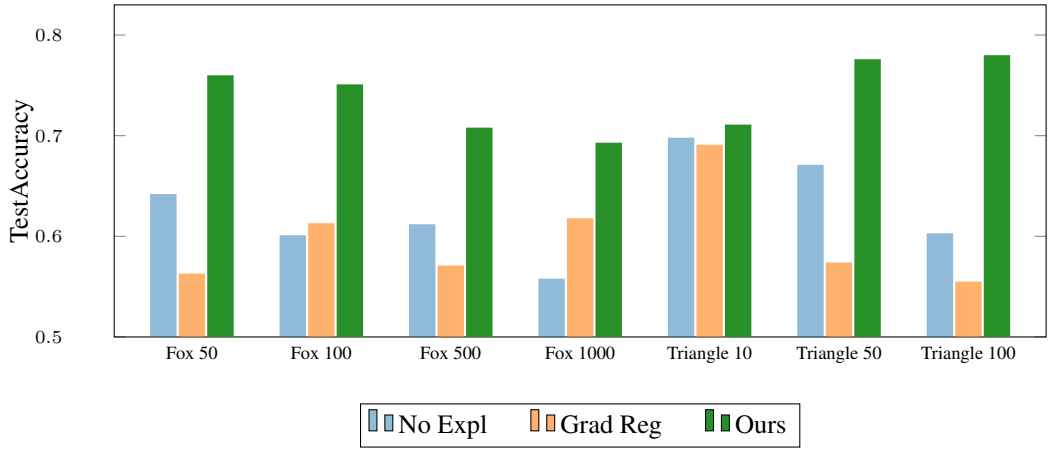}}
	\stepcounter{figureNumber}
	\fi
	\caption{Test accuracy on test sets sharing the same true reasons when models are trained on spurious training set. Our models always do better, suggesting they have learned more of the true reasons.}
	\label{fig:spurious_same_reasons}
	\end{center}
	\vskip -0.2in
\end{figure*}

We present some additional empirical results in this section.
\begin{itemize}
	\item \cref{table:std} presents model performance at convergence under different training settings with standard deviation.
	\item \cref{fig:spurious_same_reasons} presents test accuracy on test sets with the same true reasons when models are trained on spurious training set.
	\item \cref{table:spurious_training} presents model performance on clean test sets when they are trained with spurious training sets.
\end{itemize}

\begin{table*}[!ht]
	\caption{Black numbers: Final models performance on clean test sets when they are trained under different settings. {\color{red}{Red numbers}:} Final model performance on additional datasets sharing the same reasons with the training datasets when models are trained with balanced datasets. {\color{blue}{Blue numbers}:} Final model performance on balanced testsets when the class ratio in the training set is $1 : 9$. Numbers in \textbf{bold} are best performances in each setting.}
	\label{table:std}
	\vskip 0.15in
	\begin{center}
	\begin{small}
	\begin{sc}
		\begin{tabular}{ccccc}
			\toprule
			\multicolumn{1}{c}{Dataset} &\multicolumn{1}{c}{Size} &\multicolumn{1}{c}{No expl} &\multicolumn{1}{c}{Grad Reg} &\multicolumn{1}{c}{Ours} \\ 
			\midrule
			\multirow{12}{*}{Fox vs Cat} &  \multirow{3}{*}{50}  &  $0.594 \pm 0.119$  & $0.610 \pm 0.110$ &  \pmb{$0.750 \pm 0.029$} \\ 
			& & \color{red}{$0.587 \pm 0.120$} & \color{red}{$0.616 \pm 0.122$} & \color{red}\pmb{$0.772 \pm 0.037$} \\
			& & \color{blue}{$0.552 \pm 0.086$} & \color{blue}{$0.548 \pm 0.084$} & \color{blue}\pmb{$0.607 \pm 0.065$} \\
			\cmidrule{2-5}
			& \multirow{3}{*}{100}  & $0.621 \pm 0.119$  & $0.601 \pm 0.123$ &  \pmb{$0.799 \pm 0.019$} \\ 
			& & \color{red}{$0.623 \pm 0.136$} & \color{red}{$0.606 \pm 0.133$} & \color{red}\pmb{$0.813 \pm 0.033$} \\
			& & \color{blue}{$0.559 \pm 0.094$} & \color{blue}{$0.568 \pm 0.110$} & \color{blue}\pmb{$0.635 \pm 0.069$} \\
			\cmidrule{2-5}
			& \multirow{3}{*}{500}  & $0.754 \pm 0.150$  & $0.785 \pm 0.127$ &  \pmb{$0.874 \pm 0.012$} \\ 
			& & \color{red}{$0.764 \pm 0.163$} & \color{red}{$0.795 \pm 0.134$} & \color{red}\pmb{$0.890 \pm 0.018$} \\
			& & \color{blue}{$0.725 \pm 0.142$} & \color{blue}{$0.698 \pm 0.145$} & \color{blue}\pmb{$0.806 \pm 0.038$} \\
			\cmidrule{2-5}
			& \multirow{3}{*}{1000}  & $0.837 \pm 0.116$  & $0.815 \pm 0.070$ &  \pmb{$0.903 \pm 0.017$} \\ 
			& & \color{red}{$0.846 \pm 0.122$} & \color{red}{$0.822 \pm 0.083$} & \color{red}\pmb{$0.916 \pm 0.016$} \\
			& & \color{blue}{$0.822 \pm 0.081$} & \color{blue}{$0.819 \pm 0.034$} & \color{blue}\pmb{$0.824 \pm 0.034$} \\
			\midrule
			\multirow{9}{*}{Triangle} & \multirow{3}{*}{10}  & $0.595 \pm 0.118$  & $0.576 \pm 0.124$ &  \pmb{$0.732 \pm 0.082$} \\ 
			& & \color{red}{$0.562 \pm 0.092$} & \color{red}{$0.565 \pm 0.114$} & \color{red}\pmb{$0.710 \pm 0.104$} \\
			& & \color{blue}{$0.550 \pm 0.093$} & \color{blue}{$0.570 \pm 0.120$} & \color{blue}\pmb{$0.735 \pm 0.101$} \\
			\cmidrule{2-5}
			& \multirow{3}{*}{50}  & $0.726 \pm 0.154$  & $0.649 \pm 0.134$ &  \pmb{$0.839 \pm 0.028$} \\ 
			& & \color{red}{$0.713 \pm 0.158$} & \color{red}{$0.596 \pm 0.122$} & \color{red}\pmb{$0.833 \pm 0.040$} \\
			& & \color{blue}{$0.595 \pm 0.137$} & \color{blue}{$0.594 \pm 0.124$} & \color{blue}\pmb{$0.803 \pm 0.070$} \\
			\cmidrule{2-5}
			& \multirow{3}{*}{100}  & $0.687 \pm 0.175$ & $0.643 \pm 0.148$ & \pmb{$0.874 \pm 0.018$} \\
			& & \color{red}{$0.673 \pm 0.176$} & \color{red}{$0.605 \pm 0.137$} & \color{red}\pmb{$0.855 \pm 0.034$} \\
			& & \color{blue}{$0.612 \pm 0.142$} & \color{blue}{$0.567 \pm 0.129$} & \color{blue}\pmb{$0.752 \pm 0.101$} \\
			\midrule
			\multirow{2}{*}{Bird}	& 30 & $0.627 \pm 0.128$	& $0.673 \pm 0.121$		& \pmb{$0.777 \pm 0.044$} \\ 
			\cmidrule{2-5}
  			& 	60	&	$0.684 \pm 0.186$	& $0.749 \pm 0.177$		& \pmb{$0.860 \pm 0.079$} \\ 
  			\bottomrule
		\end{tabular}
	\end{sc}
	\end{small}
	\end{center}
	\vskip -0.1in
\end{table*}

\begin{table*}[!ht]
	\caption{Black numbers: Models performance on clean test sets when they are trained under different settings on datasets with spurious features. {\color{red}{Red numbers}:} Model performance on additional datasets sharing the same reasons with the training datasets. Results show that models trained with our proposed setup perform the best across all three datasets and different dataset sizes, which means our models learn the most from the given reasons in the presence of strong spurious correlations.}
	\label{table:spurious_training}
	\vskip 0.15in
	\begin{center}
	\begin{small}
	\begin{sc}
			\begin{tabular}{ccccc}
			\toprule
					\multicolumn{1}{c}{Spurious dataset} &\multicolumn{1}{c}{Size} &\multicolumn{1}{c}{No expl} &\multicolumn{1}{c}{Grad Reg} &\multicolumn{1}{c}{Ours} \\ 
					\midrule
					\multirow{8}{*}{Fox vs Cat} &  \multirow{2}{*}{50}  &  0.620  & 0.573 &  \textbf{0.729} \\ 
					& & \color{red}{0.642} & \color{red}{0.563} & \color{red}{\textbf{0.760}} \\
					& \multirow{2}{*}{100}  &  0.570  & 0.601                          &  \textbf{0.727} \\ 
					& & \color{red}{0.601} & \color{red}{0.613} & \color{red}{\textbf{0.751}} \\
		            &  \multirow{2}{*}{500}  &  0.611  & 0.561  &\textbf{0.698} \\
					& & \color{red}{0.612} & \color{red}{0.571} & \color{red}{\textbf{0.708}} \\	                               
					& \multirow{2}{*}{1000}  &  0.601  & 0.605  &  \textbf{0.728} \\ 
					& & \color{red}{0.558} & \color{red}{0.618} & \color{red}{\textbf{0.693}} \\ 
					\midrule
				    \multirow{6}{*}{Triangle} & \multirow{2}{*}{10} & 0.697 & 0.677 & \textbf{0.726} \\
				    & & \color{red}{0.698} &\color{red}{ 0.691} &\color{red}{\textbf{ 0.711}} \\
		 		    &  \multirow{2}{*}{50} &  0.694  & 0.600     &\textbf{0.796} \\ 
		 		    & & \color{red}{0.671} &\color{red}{ 0.574} &\color{red}{\textbf{ 0.776}} \\
				    &  \multirow{2}{*}{100}  &  0.606  & 0.575     &\textbf{0.784}  \\
				    & & \color{red}{0.603} &\color{red}{ 0.555} &\color{red}{\textbf{ 0.780}} \\
				    \midrule
					Bird	   & 	60	&	0.698	& 0.733		& \textbf{0.878} \\ 
			\bottomrule		
			\end{tabular}
		\end{sc}
		\end{small}
		\end{center}
		\vskip -0.1in
\end{table*}

\end{document}